\documentclass[10pt,twocolumn,letterpaper]{article}

\usepackage{iccv}
\usepackage{times}
\usepackage{epsfig}
\usepackage{graphicx}
\usepackage{amsmath}
\usepackage{amssymb}

\usepackage{colortbl}
\usepackage[dvipsnames, svgnames, x11names]{xcolor}

\usepackage{booktabs} 
\usepackage{float}
\usepackage{verbatimbox}
\usepackage{scalerel}
\usepackage{makecell}

\usepackage{bbding} 

\usepackage{rotating}         

\usepackage{multirow}

\usepackage{amssymb}
\usepackage{pifont}
\usepackage[pagebackref=true,breaklinks=true,letterpaper=true,colorlinks,bookmarks=false]{hyperref}
\usepackage[inline,shortlabels]{enumitem}
\usepackage{tabto}
\iccvfinalcopy 

\def\iccvPaperID{5762} 

\ificcvfinal\fi

\begin{document}
\newcommand*\samethanks[1][\value{footnote}]{\footnotemark[#1]}
\title{FBNet: Feature Balance Network for Urban-Scene Segmentation}

\author{
Lei Gan\thanks{These authors contributed equally} \textsuperscript{\rm 1}, Huabin Huang\samethanks[1] \textsuperscript{\rm 2}, Banghuai Li \textsuperscript{\rm 2}, Ye Yuan \textsuperscript{\rm 2} \\
\textsuperscript{\rm 1} Beihang University
\textsuperscript{\rm 2} MEGVII Technology \\
ganlei@buaa.edu.cn,\{huanghuabin, libanghuai, yuanye\}@megvii.com
}

\maketitle
\ificcvfinal\thispagestyle{empty}\fi

\begin{abstract}

Image segmentation in the urban scene has recently attracted much attention due to its success in autonomous driving systems. However, the poor performance of concerned foreground targets, e.g., traffic lights and poles, still limits its further practical applications. In urban scenes, foreground targets are always concealed in their surrounding stuff because of the special camera position and 3D perspective projection. What's worse, it exacerbates the unbalance between foreground and background classes in high-level features due to the continuous expansion of the reception field. We call it Feature Camouflage. In this paper, we present a novel add-on module, named \textbf{F}eature \textbf{B}alance \textbf{Net}work (\textbf{FBNet}), to eliminate the feature camouflage in urban-scene segmentation. FBNet consists of two key components, i.e., Block-wise BCE(BwBCE) and Dual Feature Modulator(DFM). BwBCE serves as an auxiliary loss to ensure uniform gradients for foreground classes and their surroundings during backpropagation. At the same time, DFM intends to enhance the deep representation of foreground classes in high-level features adaptively under the supervision of BwBCE. These two modules facilitate each other as a whole to ease feature camouflage effectively. Our proposed method achieves a new state-of-the-art segmentation performance on two challenging urban-scene benchmarks, i.e., Cityscapes and BDD100K. Code will be released for reproduction.

\end{abstract}

\section{Introduction}\label{introduction}
Semantic image segmentation, a fundamental task in computer vision, is widely employed in basic urban-scene understanding scenarios such as autonomous driving, image editing, \textit{etc}. With the recent development of convolution neural network-based methods \cite{fcn, resnet, cnn1, cnn2, cnn3, cnn4}, urban-scene segmentation has achieved great improvement. However, almost all these methods ignore the data distribution characteristics of the urban scene and treat it as a common segmentation task.  The performance of many important targets is still limited, and the reason hasn't been fully exploited, which restricts its further application.

\begin{figure}
\centering
\centering\includegraphics[width=8cm]{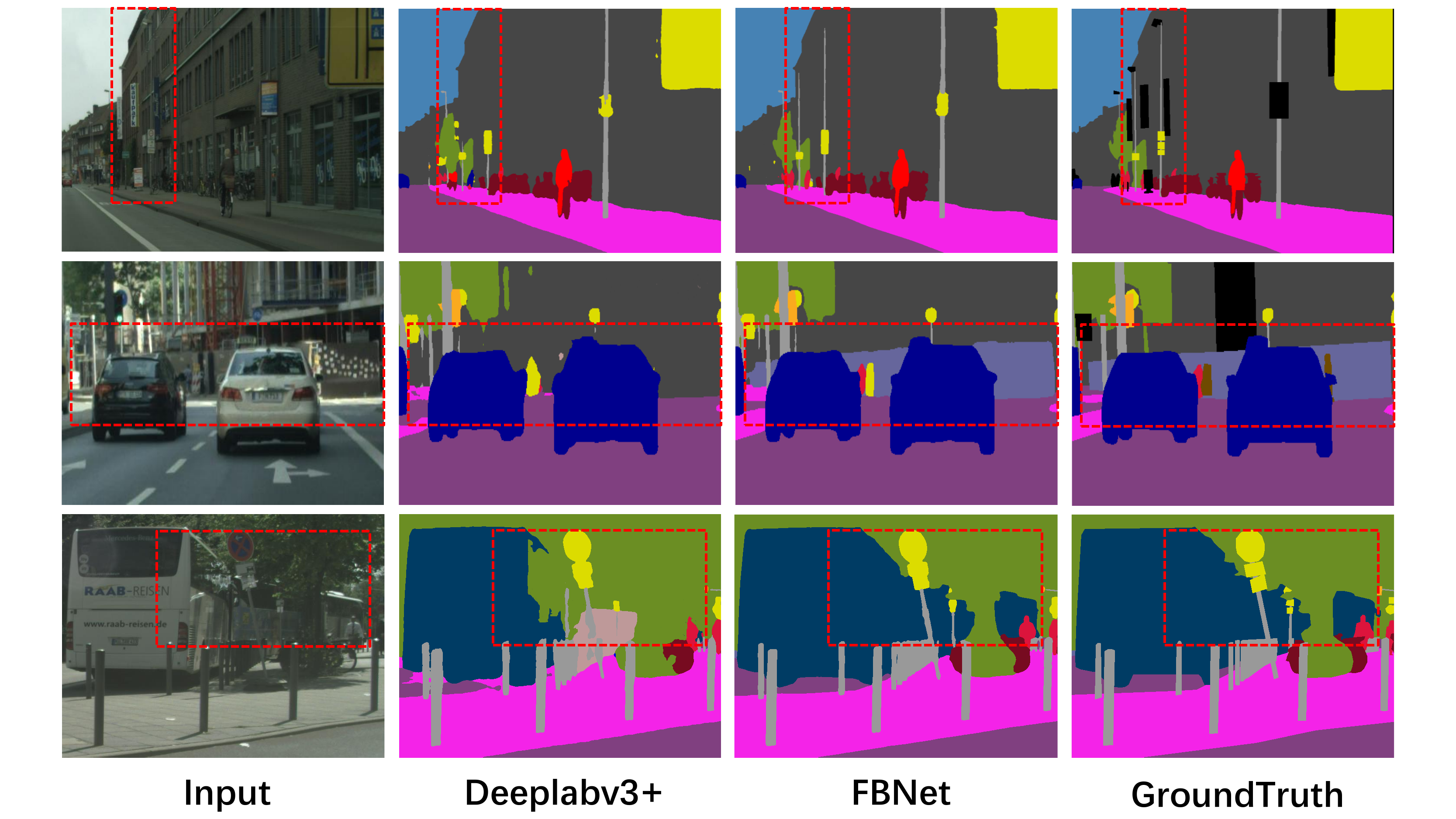}
\caption{Segmentation results based on DeepLabv3+ \cite{deeplabv3plus}.  Obvious improvements over DeepLabv3+ are marked in \textcolor{red}{red} dashed boxes. Our proposed FBNet distinguishes poles, wall, and bus pixels  from their surrounding effectively.}
\label{improvemnet_vision}
\vspace{-0.5cm}  
\end{figure}

\begin{figure}
\centering
\centering\includegraphics[width=8cm]{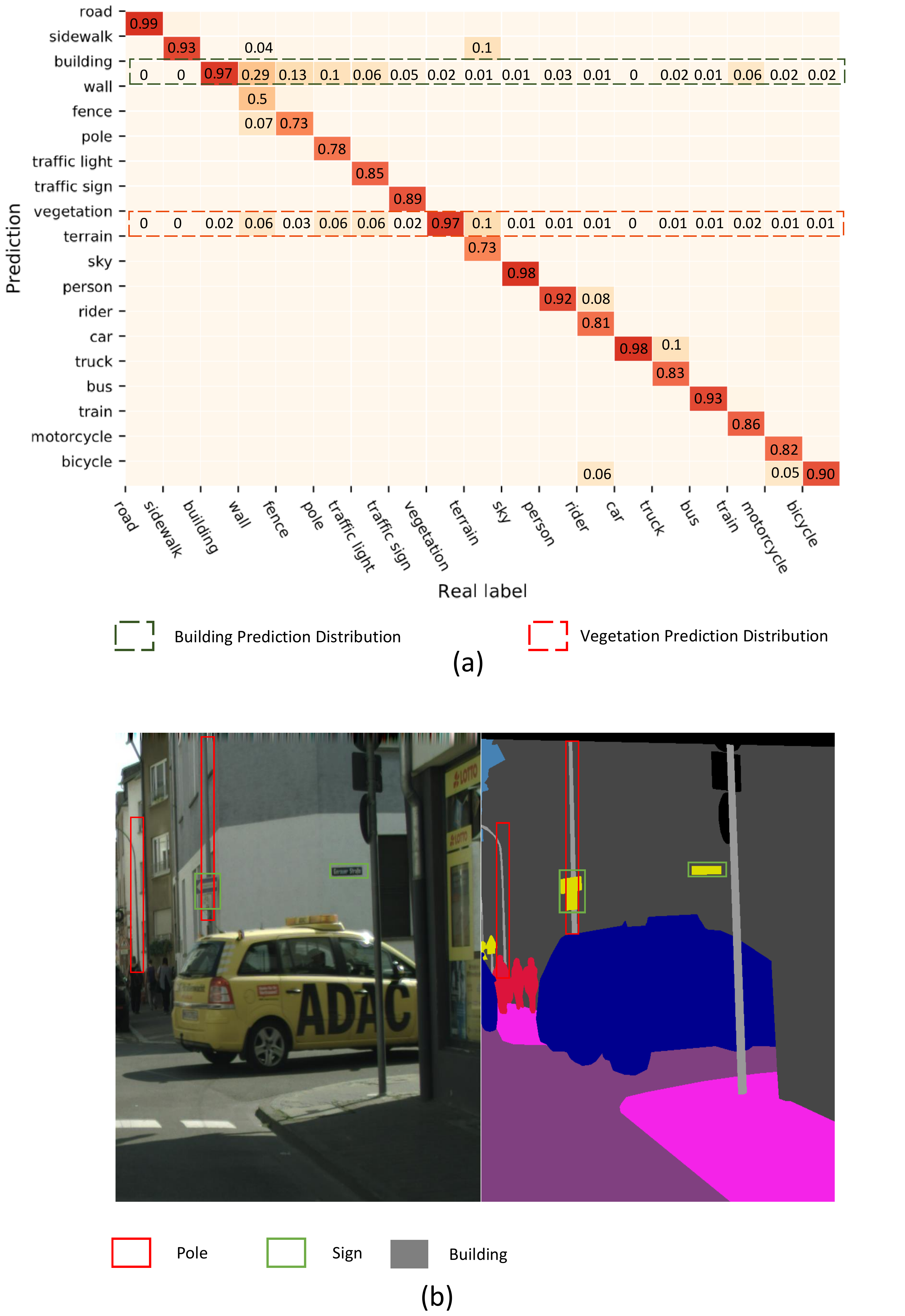}
\caption{Observations on Cityscapes~\cite{cityscapes} \textit{validation } dataset. (a) Confusion matrix of DeepLabv3+~\cite{deeplabv3plus}. We can find that many foreground target pixels are misclassified as buildings or vegetation, e.g. $\textbf{29\%}$ wall and $\textbf{13\%}$ fence pixels are misclassified as  buildings. (b) Examples of foreground classes, which are highlighted in \textcolor{red}{red} and \textcolor{green}{green} dashed boxes. 
}
\vspace{-0.5cm}  
\label{observation}
\end{figure}

\begin{figure*}
\centering\includegraphics[width=12cm]{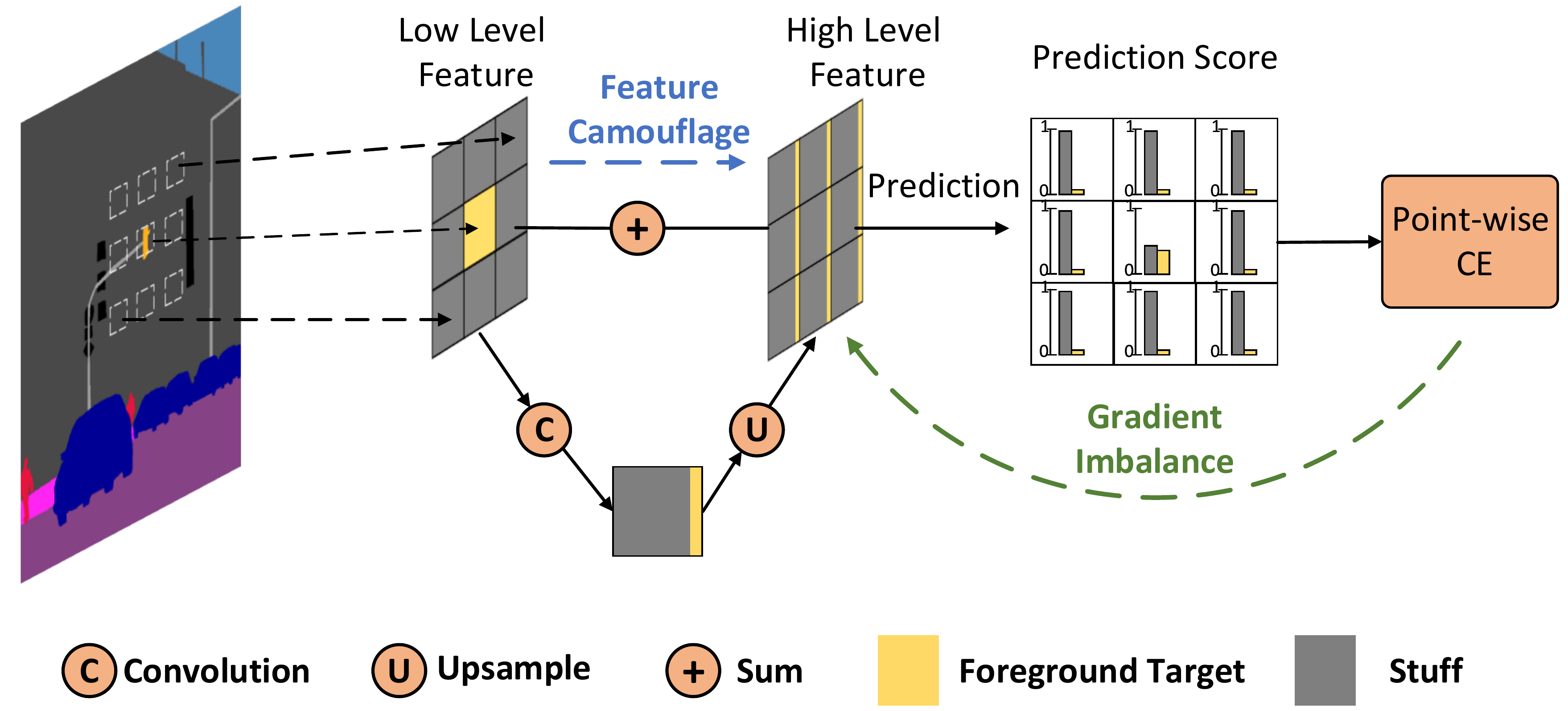}
\caption{Illustration of feature camouflage. High-level features are always achieved by a series of reception field expansion operations like convolution and dilated convolution,  which leads to the continuous dilution of foreground targets' features by the surrounding stuff. For example, the feature of \textit{lamp} is diluted by its surrounding background stuff \textit{building} in this figure. Thus, the \textit{lamp} is under-represented and results in poor performance. We call this phenomenon  \textit{Feature Camouflage}.}
\label{feature_imbalance}
\vspace{-0.5cm}  
\end{figure*}

As shown in Figure \ref{observation} (b), foreground targets like lamb poles are always concealed by surrounding background stuff like buildings due to the camera's special position and 3D perspective projection. Consequently, these foreground targets usually tend to be misclassified as their surrounding stuff by many state-of-the-art(SOTA) semantic segmentation works. We take DeepLabv3+~\cite{deeplabv3plus} as an example and obtain its confusion matrix between each class pair on Cityscapes~\cite{cityscapes} \textit{validation} dataset in Figure \ref{observation} (a). We can find that about $29\%$ wall and $13\%$ fence pixels are misclassified as buildings. We delve into this phenomenon and ascribe it to two reasons. First, background stuff\footnote{We define road, sidewalk, building, sky and vegetation as the background stuff in this paper, because all of them tends to occupy main area of the image and is widely adopted as background in practical urban-scene applications~\cite{cityscapes,bdd100k}, and the other classes we call foreground classes.} always occupy much more pixels than foreground targets in urban-scene images. Thus, foreground classes will contribute less to backpropagated gradients during model optimization, which results in a model bias towards background stuff classes. It is known widely as \textit{Class Imbalance}(\textit{Gradient Imbalance}) problem. Second, deep convolutional neural networks in semantic segmentation rely on high-level discriminative features to distinguish each class accurately. However, high-level features are always achieved by a series of reception field expansion operations like convolution and dilated convolution,  which leads to the continuous dilution of foreground target features by the surrounding background stuff. Finally, it exacerbates the unbalance between foreground and background classes in high-level feature maps. We call it \textbf{\textit{Feature Camouflage}}. Figure \ref{feature_imbalance} illustrates this issue intuitively. Several attempts \cite{deeplabv3plus, fpn, fusion1, hrnet, fusion1, fusion2} focus on fusing low-level features with high-level features may ease the problem implicitly, but high-level features are more discriminative\cite{high_1, high_2, high_3, high_4} and still dominate the final classification. 

Although many methods like loss re-weighting~\cite{re-weight1, re-weight2, re-weight3, re-weight5} and data over-sampling~\cite{urban_scene_25, re-sample1, re-sample2, re-sample3, re-sample4} 
have been proposed to relieve the class imbalance. We argue that these methods have a very limited impact on the feature camouflage problem. They only utilize the pixel number statistics of images or labels to balance the overall performance, while the essential analysis on the high-level features is seldom exploited.  In this paper, we propose a novel Feature Balance Network (FBNet) as an add-on module to solve the feature camouflage problem by exploring the spatial relationship between foreground classes and background stuff classes. FBNet aims to balance the high-level feature explicitly and highlights more foreground targets from their surrounding stuff. It consists of two key components, i.e., Block-wise Binary Cross-Entropy (BwBCE) and Dual Feature Modulator(DFM).




BwBCE serves as an auxiliary loss to enhance the representation of foreground classes in the deep layers to ease the feature camouflage problem. As we all know, point-wise cross-entropy is widely adopted in the past semantic segmentation works to achieve the analogous purpose. However, point-wise cross-entropy only concentrates on the single pixel to classify it into the correct class. Obviously, if some class occupies the most pixels, it will dominate the model optimization due to the cumulative effect, while the scarce target class will be suppressed correspondingly. In contrast, our proposed BwBCE is designed from a gradient balanced perspective. BwBCE treats all pixels belonging to the same class as a whole to ensure uniform gradient for the foreground class and its surroundings during backpropagation. It highlights the existence of foreground class in the high-level features and obtains a more robust feature representation for foreground classes.

According to the above discussion, BwBCE tends to cause potential conflict with the final segmentation loss, which is a typical point-wise cross-entropy loss, if we combine them together to supervise the model training directly. For this reason, we propose a flexible yet effective module named DFM to make them work in a more proper way. DFM consists of two main components, i.e., channel sensor and spatial sensor. These two modules attempt to strengthen the feature representation of foreground classes from the channel and spatial dimension respectively under the full supervision of BwBCE. Then, fusing the output feature of DFM with the original one will solve the feature camouflage problem effectively and boost the overall performance.


It is worth mentioning that our proposed FBNet is orthogonal with the existing SOTA methods. Methods equipped with our approach can achieve further improvements with negligible cost, verified in Section \ref{ablation}. Apart from this, we conduct extensive experiments in Section \ref{experiment} to prove each component's effectiveness in FBNet. Superior performance on two challenging urban-scene segmentation benchmarks, i.e., Cityscapes~\cite{cityscapes} and BDD100K~\cite{bdd100k}, shows the effectiveness and robustness of our proposed method as well. In summary, the contributions of this paper are three-fold:
\begin{itemize}
\item We delve into the poor performance of foreground classes in the urban-scene segmentation and attribute it to feature camouflage empirically, which is seldom exploited in the past works.
\item We propose a novel module named Feature Balance Network(FBNet), which consists of two key components, i.e., Block-wise Binary Cross-Entropy (BwBCE) and Dual Feature Modulator(DFM), to ease the feature camouflage effectively.
\item Our proposed method can achieve superior performance over all the SOTA methods on two challenging urban-scene benchmarks~\cite{cityscapes,bdd100k}, especially on foreground categories.
 \end{itemize}

\section{Related Work}
\textbf{Semantic Segmentation}
Semantic Segmentation is a fundamental visual task aiming at setting class label for every pixel of the image. Maintaining the resolution of a feature map while capturing discriminative high-level  features is vital important for high performance of semantic segmentation. Usually, high-level features are extracted by some continuous downsample convolutions and spatial pooling layers, but the resolution gets coarser in the procedure because of the growing reception field. Several researches \cite{fcn, cars_33} relieve this problem by leveraging deconvolution for learnable upsampling from low-resolution features. Some more intuitive and effective method  \cite{SegNet, unet, cars_24, deeplabv3plus} hire skip-connections to maintain high-resolution and discriminative fused feature to recover the object boundaries. Another prevalent method is atrous convolution \cite{d_33, DilatedFCN}, which increases the receptive field size while keeping resolution without increasing the number of parameters, and it is widely adopted in recent semantic segmentation networks \cite{DeepLabV3, deeplabv3plus, denseaspp, zhang2018context, uniformsample}. Additionally, ASPP \cite{DeepLabV3} and pyramid pooling modules \cite{pspnet} address such challenges caused by diverse scales of objects.

\textbf{ Urban-scene Exploitation}
In the field of urban-scene parsing, several studies exploit the urban-scene images characteristics. In general, the scale of targets significantly vary in the urban-scene images. FoveaNet \cite{cars_23} localizes a “fovea region”, where the small scale objects are crowded, and performs scale normalization to address heterogeneous object scales.  Another recent approach \cite{uniformsample} exploits the fact that the urban-scene images have continuous video frame sequences and proposes the data augmentation technique based on a video prediction model to create future frames and their labels. HANet \cite{HANet} emphasizes informative features or classes selectively according to the vertical position of a pixel. Another approach \cite{cars_10} separates an urban-scene image into several spatial regions and conducts domain adaptation on the pixel-level features from the same spatial region.

\textbf{Class Imbalance}
In urban-scene images, few categories occupy most pixels which results minority classes can not get enough gradient during training strategy. Thus, the performance of minority classes is often not as good as majority classes. This is the widely studied class imbalance problem. Many approaches have been proposed to deal with class imbalance. Some use re-sampling strategies \cite{re-sample1, re-sample2, re-sample3, re-sample4} to re-balance data distribution. In their basic versions, the dataset is balanced by increasing the number of objects from “minority” classes, respectively. But oversampling methods tend to over-fit due to the inclusion of duplicate data. Re-weighting \cite{re-weight1, re-weight2, re-weight3} is another widely used approach. It assigns each category different weights according to the points number to simply increase minority classes gradient. All these methods only care about points number of different targets but do not take into account the positional relationship between them, which limits their ability to refine weak target segmentation.

\begin{figure}
\centering\includegraphics[width=7cm]{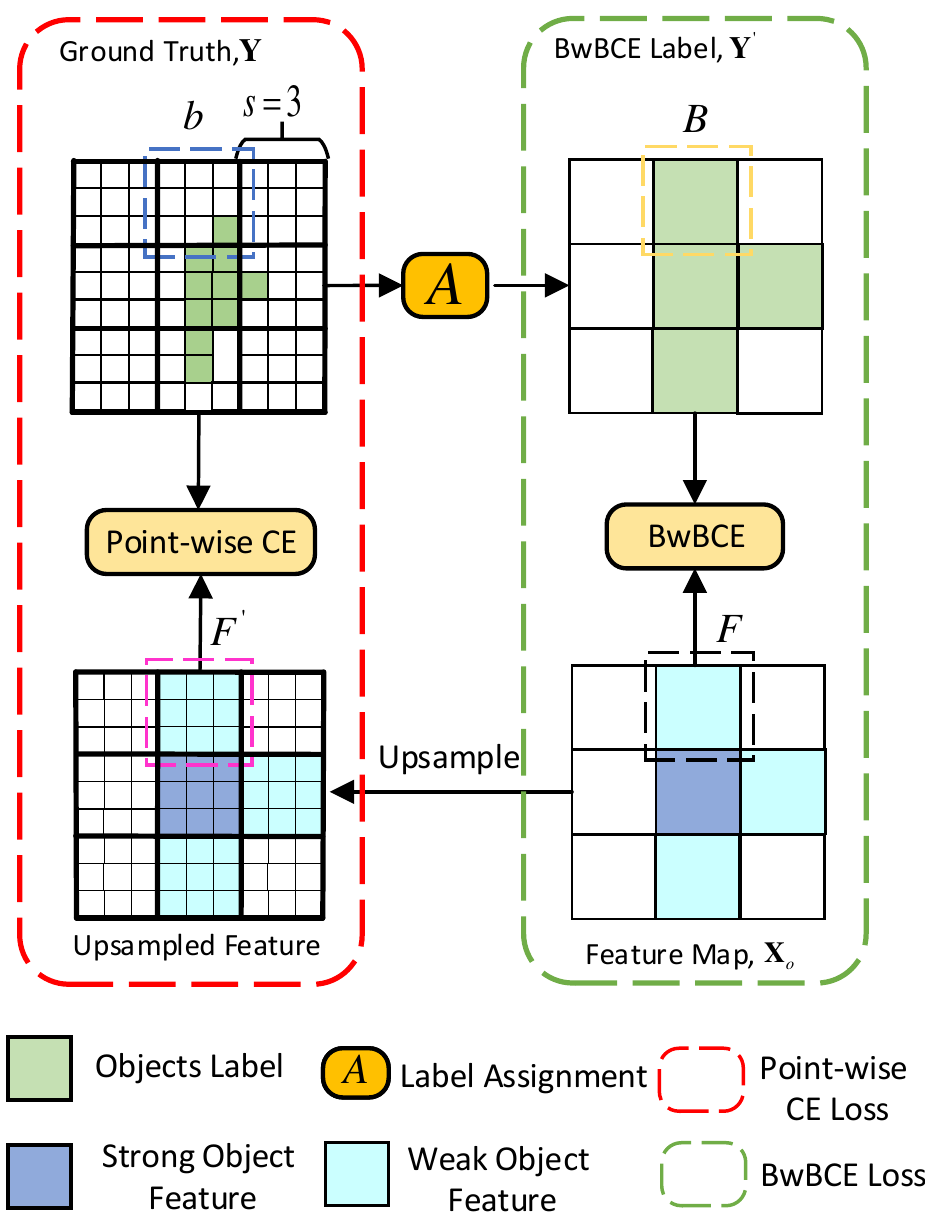}
\caption{Illustration of difference between BwBCE and Point-wise cross-entropy. $s$ is the stride of high-level feature maps, $Y$ is the the ground-truth label mask. $A$ means the label assignment process to convert $Y$ into $Y^{'}$ to match the output size of $X_{o}$. 
$F$ in high level feature map is weak foreground target feature because of feature camouflage. The feature points in dash box $F^{'}$ are upsampled from $F$.
}
\label{BwBCE_G}
\end{figure}

\begin{figure*}
\centering\includegraphics[width=12cm]{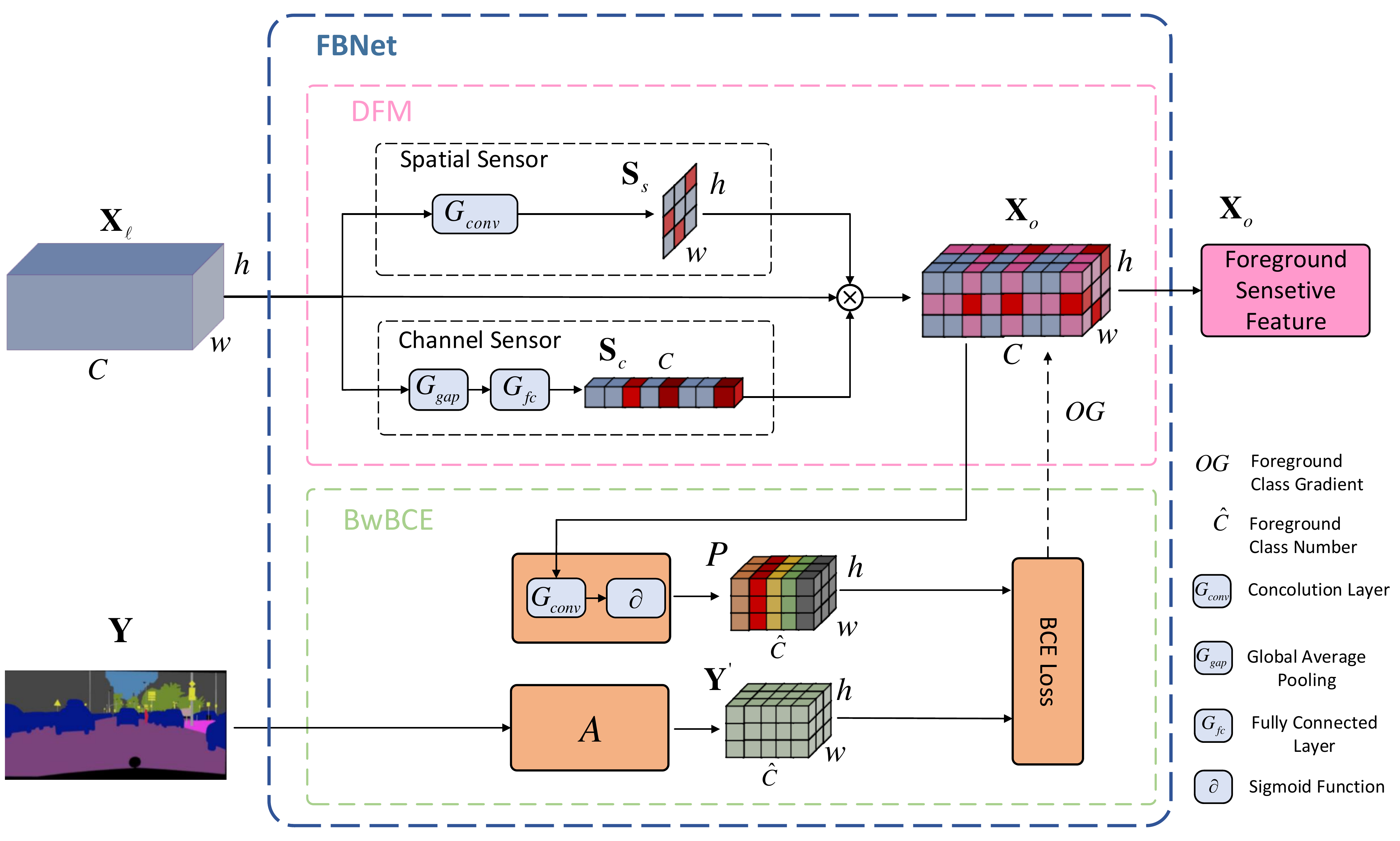}
\caption{Architecture of our proposed FBNet. FBNet is composed of two main parts, DFM and BwBCE.  DFM applies spatial and channel sensor on the original feature $\textbf{X}_{\ell}$ to get spatial weight $\textbf{S}_{s}$ and channel weight $\textbf{S}_c$ respectively. Then, output features $\textbf{X}_{o}$ can be obtained by fusing $\textbf{S}_{s}$, $\textbf{S}_{c}$ with $\textbf{X}_{\ell}$. BwBCE compels $\textbf{X}_{o}$ to preserve more foreground target features during the training procedure. The key factor label generator $A$ generates the foreground target label $\textbf{Y}^{'}$ from original label $\textbf{Y}$. Details can be found in Section \ref{FM}.}
\label{FBNet}
\end{figure*}

\section{Proposed Method}\label{Method}
In this paper, we propose a novel method named Feature Balance Network (FBNet)  from feature manipulation perspective. It consists of two key components, i.e., Block-wise Binary Cross-Entropy(BwBCE) Loss and Dual Feature Modulator(DFM). FBNet intends to enhance foreground high-level features to relieve the feature camouflage problem. An overview of our pipeline can be found in Figure \ref{FBNet}. In the following section, we will illustrate BwBCE and DFM respectively, followed by the details in training objective and network architecture.

\subsection{Block-wise Binary Cross Entropy Loss}\label{Loss}
FBNet intends to strengthen the deep representation of foreground classes in high-level features. Thus, the fundamental factor is to perceive where to perform the optimization. In this section, we design a simple but effective auxiliary loss named Block-wise Binary Cross-Entropy (BwBCE) loss to achieve the goal.

In fact, introducing auxiliary loss for robust feature learning in deep layers is a common strategy in semantic segmentation~\cite{MMSeg}. However, almost all these methods \cite{pspnet, zhang2018context, detr, BiseNet} adopt the point-wise cross-entropy as the auxiliary loss, which has no difference with the final classification loss. Point-wise, cross-entropy has the inherent drawback towards the feature camouflage problem. As shown in Figure \ref{BwBCE_G}, high-level feature maps are always upsampled to the size of the label mask, then each corresponding pixel pair in the feature map and label mask is used to calculate the cross-entropy loss. Assuming $F$ $\in\mathbb{R}^{1 \times 1}$ is one single pixel in the high-level feature map (bottom-right corner in Figure \ref{BwBCE_G}), $F$ will be 3x upsampled to $F^{'}$ $\in\mathbb{R}^{3 \times 3}$ to match the size of label mask. If only one pixel in $F^{'}$ is marked as foreground class in the label mask (top-left corner in Figure \ref{BwBCE_G}), merely 1/9 of the gradient in $F^{'}$ contributes to the foreground class optimization during backpropagation, which harms the foreground class feature presentation due to the gradient dilution.


On the contrary, our proposed BwBCE is designed to ease feature camouflage explicitly from a gradient perspective. As shown in Figure \ref{FBNet}, $P$ $\in\mathbb{R}^{C \times h \times w}$ is the prediction result, $\hat{C}$ is the number of foreground classes, $Y$ $\in\mathbb{R}^{H \times W}$ is the ground-truth label mask. $A$ means the label assignment to convert $Y$ into $Y^{'}$ $\in\mathbb{R}^{C \times h \times w}$, and $Y^{'}$ is in a multi-hot format which serves as the direct supervision of $P$. Formally, $A$ is defined as follows:



\begin{small}
\begin{equation}
\begin{aligned}
Y^{'} = A(Y)
\end{aligned}
\end{equation}
\end{small}
and
\begin{small}
\begin{equation}
\begin{aligned}
Y^{'}_{c,i,j} =\varepsilon(\sum_{k=s\cdot i}^{s\cdot(i+1)}\sum_{l=s\cdot j}^{s\cdot(j+1)}{Y_{k,l}=c})
\end{aligned}
\end{equation}
\end{small}
\noindent where $c\in[ 0, \hat{C}), i \in [0,h), j \in [0, w)$,  $s = \frac{H}{h}$ or $s = \frac{W}{w}$ is the stride of  $P$, $\varepsilon(x)=1$ if $x\geq 0$ and 0 otherwise. In a nutshell, if any pixel of class $c$ falls on  block $b\in Y^{s \times s}$, the corresponding position of multi-hot vector in $Y^{'}$ will be set 1 else 0, as shown in Figure \ref{BwBCE_G}. In addition, $P$ is defined  as follows:
\begin{equation}
P = \sigma(G_{conv}(X_{o}))
\end{equation}


\noindent$\sigma$ stands for the \textit{sigmoid} function, $G_{conv}$ is a typical convolution operation, and $X_{o}$ is the input high-level features. Ultimately, our proposed BwBCE loss can be described as:
\begin{small}
\begin{equation}
\begin{aligned}
{L}_{BwBCE}(P,Y^{'})=\sum_{c=0}^{\hat{C}}\sum_{i=0}^{h}\sum_{j=0}^{w}{Y^{'}_{c,i,j}\cdot log P_{c,i,j}}\\+(1-Y^{'}_{c,i,j})\cdot log(1-P_{c,i,j})
\end{aligned}
\end{equation}
\end{small}

Comparing our BwBCE with point-wise cross-entropy, the main difference lies in that  BwBCE ensures uniform gradient for foreground class and its surroundings during backpropagation even if huge pixel imbalance exists between them, while point-wise cross-entropy tends to result in serious gradient dilution for scarce foreground class. What's more, even loss re-weighting strategies is adopted to point-wise cross-entropy is incapable of solving this problem. 
Thus, BwBCE is more suitable as the auxiliary loss for more balance feature learning.





\subsection{Dual Feature Modulator}\label{FM}
Generally speaking, point-wise cross-entropy loss is widely adopted for the final classification to distinguish each pixel in the semantic segmentation. However, BwBCE just focuses on enhancing the representation of scarce foreground class in high-level features, which goes against the fine-grained classification to some extend as we have discussed in Section \ref{Loss}. It may tend to arouse inconsistent optimization objectives when we combine them as a whole to supervise the same features blindly.


To avoid the potential conflict, we design a flexible yet effective module named \textbf{\textit{Dual Feature Modulator}(DFM)}. DFM transforms the given features under the full supervision of BwBCE loss, which is independent of the point-wise cross-entropy for the final classification. Thereby, the output of DFM achieves better feature representation for the foreground class.  Mixing the transformed features with the original ones will contribute to the final performance. Several visualization examples shown in Figure \ref{attention_vision} can illustrate the effectiveness of DFM intuitively.

As shown in Figure \ref{FBNet}, DFM is designed on both channel and spatial dimensions.  We summary them as \textbf{Channel Sensor} and \textbf{Spatial Sensor} respectively. The channel sensor plays an important role in identifying which channel is more important for foreground targets' segmentation, while the spatial sensor intends to perceive the specific location of the foreground targets. Their outputs, $\textbf{S}_{s}$ and $\textbf{S}_{c}$, are defined as follows:

\begin{equation}\label{functionAs}
\textbf{S}_{s} = \sigma(G_{conv} (\textbf{X}_{\ell}) )
\end{equation}
\begin{equation}\label{functionAc}
\textbf{S}_{c} = \sigma(G_{fc}(G_{pool} (\textbf{X}_{\ell}) ))
\end{equation}
$\text{G}_{conv}$ is a $1 \times 1$ convolution operation, $G_{pool}$ is a global average pooling operation ,and $G_{fc}$ is a fully connected layer. $\sigma$ is the sigmoid function to normalize $\textbf{S}_{s}$ and  $\textbf{S}_{c}$ between 0 and 1.
Eventually, the output feature map $\textbf{X}_{o}$ can be described as:
\begin{equation}\label{...}
  \textbf{X}_{o} =  \textbf{X}_{\ell}\otimes G_{bro} (\textbf{S}_{s})\otimes G_{bro}(\textbf{S}_{c})
  \end{equation}
$\otimes$ means points-wise multiplication. $G_{bro}$ is a broadcast operation to ensure $\textbf{S}_{s}$ and $\textbf{S}_{c}$ have the same shape with $\textbf{X}_{l}$. As a result, $\textbf{X}_{o}$ is supposed to benefit to foreground classes in feature representation. In the end, we fuse $\textbf{X}_{o}$ with $\textbf{X}_{l}$ to modulate the deep representation adaptively during the model training.

It is worth noting that BwBCE and DFM are designed as a whole to facilitate each other. Extensive experiments in Section \ref{ablation} can well verify the relationship between them.

\begin{table*}
\centering
\small
\setlength{\tabcolsep}{0.4mm}{
\begin{tabular}{c|c|ccccccc >{\columncolor{LightBlue1}} c >{\columncolor{LightBlue1}} c >{\columncolor{LightBlue1}} c >{\columncolor{LightBlue1}} c >{\columncolor{LightBlue1}} c >{\columncolor{LightBlue1}} c >{\columncolor{LightBlue1}} c >{\columncolor{LightBlue1}} c >{\columncolor{LightBlue1}} c >{\columncolor{LightBlue1}} c >{\columncolor{LightBlue1}} c >{\columncolor{LightBlue1}} c >{\columncolor{LightBlue1}}c >{\columncolor{LightBlue1}} c} 
\toprule  
Method & Backbone & mIoU & f-mIoU & road & swalk & build. & veg & sky & wall & fence & pole & tligh. & tsign. & terr. & pers. & rider & car & truck & bus & train & mcyc & bcyc \\
\hline
 FCN & \multirow{2}{*}{ResNet-50} & 76.3 & 70.5 & 98.0 & 84.4 & 92.4 & \textbf{92.7} & 94.6 & 47.0 & 59.0 & 66.7 & 73.4 & 80.3 & 61.2 & 82.8 & 63.0 & 94.7 & 62.3 & 83.0 & 71.0 & 63.5 & 78.9  \\
+Ours & ~ & \textbf{78.5} & \textbf{73.4} & \textbf{98.2} & \textbf{85.6} & \textbf{92.9} & \textbf{92.7} & \textbf{94.8} &
\textbf{55.4} & \textbf{61.1} & \textbf{67.5} & \textbf{74.4} & \textbf{81.3} &  \textbf{62.7} & \textbf{83.5} & \textbf{63.9} & \textbf{95.5} &\textbf{72.8} & \textbf{89.1} & \textbf{76.2} & \textbf{64.9} & \textbf{79.6}  \\
\hline
FCN & \multirow{2}{*}{ResNet-101} & 77.8 & 72.4 & \textbf{98.4} & \textbf{86.6} & 93.1 & 92.8 & 95.1 & 56.0 & 62.5 & 68.9 & 74.1 & 82.1 & 64.6 & 83.9 & 65.3 & 95.0 & 62.6 & 84.2 & 66.9 & 67.5 & 79.4  \\

+Ours & ~ & \textbf{80.7} & \textbf{76.2} & 98.2 & 86.0 & \textbf{93.5} & \textbf{93.1} & \textbf{95.3} & \textbf{62.9} & \textbf{63.7} & \textbf{69.6} & \textbf{75.0} & \textbf{82.7} & \textbf{65.2} & \textbf{84.5} & \textbf{66.7} & \textbf{95.7} & \textbf{77.4} & \textbf{89.3} & \textbf{84.5} & \textbf{69.8} & \textbf{80.1}  \\
\bottomrule 
\end{tabular}}
\caption{Fine-grained segmentation results on the Cityscapes \textit{validation} set. Dilated FCN~\cite{DilatedFCN} is adopted as the baseline to evaluate on both ResNet-50 and ResNet-101 backbones. Note that our method improves all foreground categories which are marked in blue.} 
\label{tab_fcn_baseline} 
\end{table*}

\subsection{Training Objectives}
The overall loss of our proposed method is
\begin{equation}
L = \lambda_{1} \sum_{f=1}^F{L}_{BwBCE}(\textbf{P},\textbf{$Y^{'}$}) + \lambda_{2}L_{ce}({\textbf{Z}},{\textbf{Y}}) )
\end{equation}
$L_{ce}$ is the typical cross-entropy loss for segmentation, $\textbf{Z}$ is the final classification score map, and $\textbf{Y}$ is the corresponding pixel annotations. $\lambda_{1}$ and $\lambda_{2}$ are loss weight hyper-parameters to balance the scale of the losses, which we set both 1 as default.

\subsection{Network architecture}
\begin{figure}
\centering\includegraphics[width=7cm]{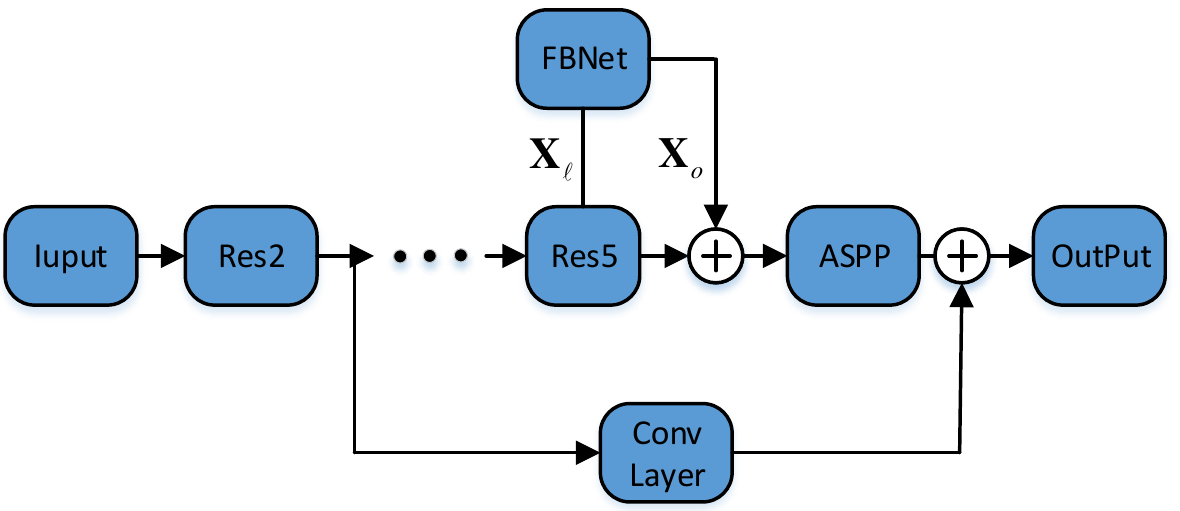}
\caption{Semantic segmentation networks incorporating FBNet. FBNet generates feature map $\text{X}_{o}$ which is rich in foreground targets feature, then original feature map $\text{X}_{\ell}$ and $\text{X}_{o}$will be fused to generate new high level feature map.}
\label{NET}
\end{figure}

Figure \ref{NET} illustrates the overall pipeline of our proposed FBNet based on the SOTA model DeepLabv3+ \cite{deeplabv3plus}. FBNet is designed as a plug-and-play module to insert into any layer from Res2 to Res5. Moreover, we conduct extensive ablation studies in Section \ref{ablation} to determine the performance gain by adding FBNet into different layers. 

\section{Experiments}\label{experiment}
In this section, we first introduce two typical urban-scene segmentation datasets, i.e., Cityscapes\cite{cityscapes} and Berkeley Deep Drive dataset (BDD100K)\cite{bdd100k}, followed by our training protocols. Then we conduct quantitative ablation studies on multiple backbones (e.g. ResNet-50~\cite{resnet} and ResNet-101~\cite{resnet}) and several different architectures(e.g. FCN \cite{fcn}, PSPNet \cite{pspnet} and DeepLabv3+ \cite{deeplabv3plus}) to demonstrate the effectiveness and wide applicability of our proposed methods.
Furthermore, we visualize the feature map generated by FBNet to verify the effect of FBNet intuitively. 

We follow standard settings~\cite{AAF, ACFNet,BiseNet, CCNet, OCR, CPNet} to measure the segmentation performance in terms of the \textbf{m}ean \textbf{I}ntersection over \textbf{U}nion (mIoU) metric. Particularly, for the performance evaluation on forground classes, we adopt the f-mIoU metric, which stands for mIoU of foreground classes.

\subsection{Datasets and training protocols}
\noindent\textbf{Cityscapes} \cite{cityscapes} is a large urban-scene dataset, which has 5000 finely annotated images and 20000 coarsely annotated images. The finely annotated data is split into 2975 training images, 500 validation images, and 1525 testing images. 
The whole dataset has 19 classes for semantic segmentation, and the image size is $ 1024 \times 2048 $ by default.

\noindent\textbf{Berkeley Deep Drive} dataset (BDD100K) \cite{bdd100k}
is a large urban-scene dataset consisting of 7,000 training and 1,000 validation images for semantic segmentation. The resolution of each image is $ 1280 \times 720 $. It's really challenging because it covers various scenes including day, night and diverse weather conditions.

\noindent\textbf{Implementation Details.}
We use SGD with momentum 0.9 and weight decay 1e-4 for optimization. The poly schedule \cite{liu2015parsenet} is used to decay the learning rate which means the initial learning rate is multiplied by $ (1-\frac{iter}{max\_iter})^{0.9} $. We set the initial learning rate as 0.01 and the power as 0.9.  We train the model for 180 epochs and on both Cityscapes and BDD100K with a batch size of 8. During model training, we perform random resizing with a scale range in [0.5, 2.0], random flipping and random cropping with a crop size of 832 $\times$ 832 on the input images. 
It is worth noting that results reported in this paper are based on a sliding window with a single scale across the whole image unless explicitly specified. In addition, we adopt a strong baseline with OHEM~\cite{OHEM}, data oversampling and loss re-weighting strategies, which are always used towards gradient imbalance as our basic model.
We conduct all experiments on a machine with  8 2080Ti GPUs, which minimizes the barriers to make all our results easy to reproduce.

\subsection{Ablation studies}\label{ablation}
\noindent\textbf{Ablation for different backbones.}
We take Dilated FCN~\cite{DilatedFCN} as our baseline model to evaluate FBNet based on two classic backbones, i.e., ResNet-50~\cite{resnet} and ResNet-101~\cite{resnet}. Detailed results on Cityscapes \textit{validation} set are listed in Table \ref{tab_fcn_baseline}. It is obvious that Dilated FCN with FBNet achieves significant improvements over corresponding baselines on both ResNet-50 and ResNet-101 backbones. For example, FBNet based on ResNet-101 can boost overall segmentation performance from $\textbf{77.8\%}$ to $\textbf{80.7\%}$ on mIoU and from $\textbf{72.4\%}$ to $\textbf{76.2\%}$ on f-mIoU, which achieves about $\textbf{2.9\%}$ and $\textbf{3.8\%}$ improvements respectively. 


\noindent\textbf{Ablation over different state-of-the-art methods.} 
Our motivation is to design FBNet as a plug-and-play module. Thus, we integrate FBNet with several strong SOTA models like PSPNet \cite{pspnet} and DeepLabv3+ \cite{deeplabv3plus} to verify the effectiveness of FBNet. We can find from Table \ref{tab_different_baseline} that FBNet achieves up to $\textbf{1.2\%}$ and $\textbf{1.7\%}$ improvements on mIoU and f-mIoU respectively. In addition, we also report the total FLOPs of each model based on $1024 \times 1024$ resolution for a more clear comparison. Our FBNet only increases 0.76 GFLOPS(around $\textbf{0.1\%}$) over PSPNet and DeepLabv3+, which indicates that FBNet can be widely adopted to any SOTA frameworks to achieve further improvements with negligible cost.

\begin{figure*}
\centering\includegraphics[width=15cm]{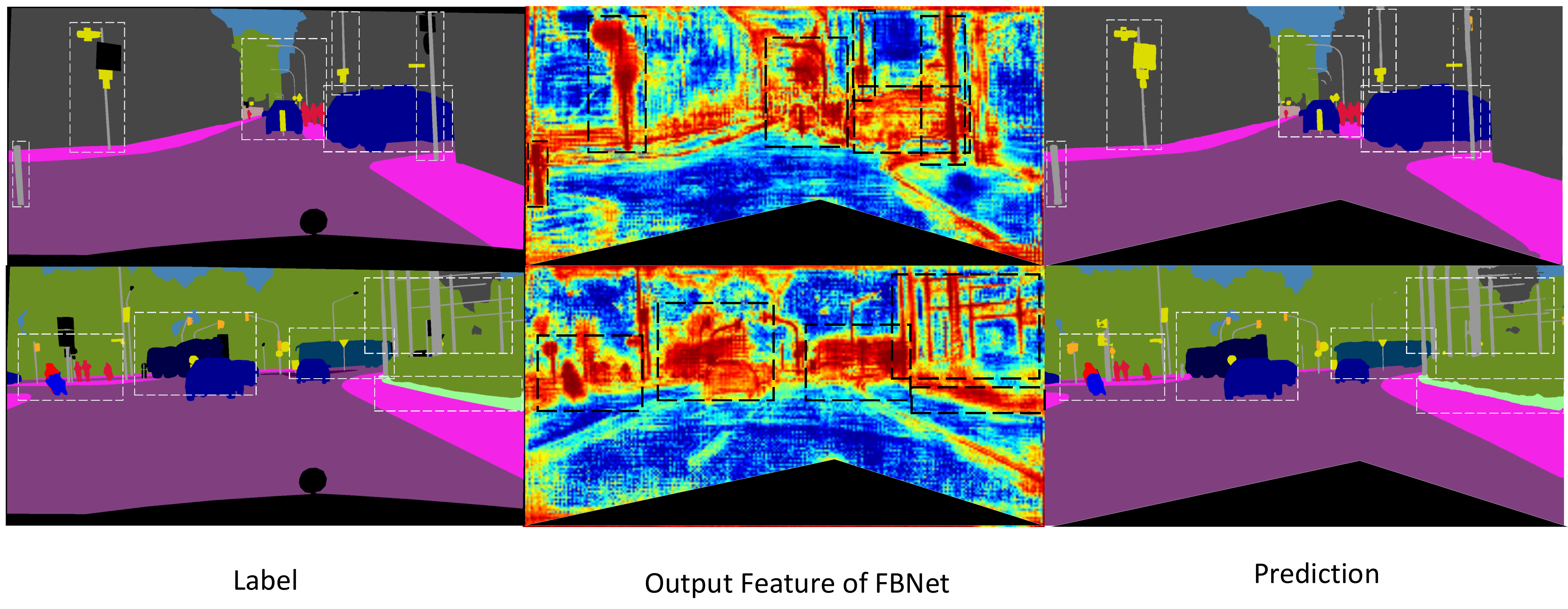}
\caption{: Visualization of FBNet output feature map. We do reduction mean along channel dimension, the feature intensity gradually increases from blue to red, for better visualization, we hide the ignored area.It can be easily noticed that the foreground targets feature are much stronger than background stuff feature. Thus adding this feature map back to the original feature map can relieve feature camouflage.}
\label{attention_vision}
\end{figure*}

 \begin{table}[H]
\centering
\setlength{\tabcolsep}{1mm}{
\begin{tabular}{l|ccccc}
\toprule 
Method & mIoU & $\Delta$ & f-mIoU & $\Delta$ &GFLOPs \\
\toprule
PSPNet~\cite{pspnet} & 79.3 & - & 74.5 & - & 540.78 \\
DeepLabv3+~\cite{deeplabv3plus} & 79.5 & - & 74.7 & - & 760.44 \\
\midrule
\small\textbf{Ours+PSPNet} & \textbf{80.5} & \textbf{1.2} \textcolor{Red}{\textbf{$\uparrow$}}  & \textbf{76.2} & \textbf{1.7} \textcolor{Red}{\textbf{$\uparrow$}} &541.54 \\
\small\textbf{Ours+DeepLabv3+} & \textbf{80.6} & \textbf{1.1} \textcolor{Red}{\textbf{$\uparrow$}} & \textbf{76.1} & \textbf{1.4} \textcolor{Red}{\textbf{$\uparrow$}} & 761.20 \\
\bottomrule 
\end{tabular}}
\caption{Ablation over different SOTA methods, where ResNet-50 serves as the backbone. The GFLOps are computed based on $ 1024 \times 1024 $ resolution.} 
\label{tab_different_baseline}
\end{table}

\noindent\textbf{Ablation for the proper layer to inject FBNet.} In this section, we conduct extensive experiments to explore the proper layer (from Res2 to Res5) to inject our FBNet. As we can see from Table \ref{tab_ab_position}, no matter where we inject FBNet, it can obtain consistent improvements and maximum is achieved when FBNet is plugged in the deepest layer. However, it is noteworthy that if we inject FBNet to all the layers or the deepest layer, they actually achieve almost the same results. In fact, it is in line with our expectations as high-level layers suffer from a more serious feature camouflage problem due to the continuous expansion of the receptive field, which exactly satisfies our assumption in Section \ref{introduction}.

\begin{table}[H]
\centering
\setlength{\tabcolsep}{1.4mm}{
 \begin{tabular}{ccccc|c|c} 
\toprule  
Method & Res2 & Res3 & Res4 & Res5  &mIoU & f-mIoU \\
\toprule
Baseline & - & -  & - & - & 77.8 & 70.6 \\
\midrule
& \checkmark  &  & & & 78.7 & 73.6 \\
& & \checkmark  &  & & 79.5 & 74.6 \\
+FBNet &  &   & \checkmark  &  & 80.1 & 75.4 \\
&  &   &   & \checkmark & 80.7 & 76.2 \\
& \checkmark   & \checkmark  & \checkmark  & \checkmark  & \textbf{80.8} & \textbf{76.3} \\
\bottomrule 
\end{tabular}}
\caption{Ablation on the proper layer to inject FBNet. Res2 $\sim$ Res5 mean the different stages in Dilated FCN~\cite{DilatedFCN} with ResNet-101 backbone. } 
\label{tab_ab_position} 
 \end{table}

\noindent\textbf{Ablation on each component of FBNet.}
As we have discussed in Section \ref{Method}, FBNet consists of two key components, i.e., BwBCE and DFM. Thus, we conduct comprehensive experiments from auxiliary loss and feature modulator perspectives to verify the superiority of our proposed components in FBNet. As shown in Table \ref{tab_ab_DFI}, even if we only adopt one of BwBCE and DFM into the baseline model, it can achieve about $\textbf{0.9\%}$ and $\textbf{1.2\%}$  improvements respectively. Apart from this, we combine the different choices of auxiliary loss and feature modulator together to clarify the impact of each component. We can find that although point-wise cross-entropy and $1 \times 1 $ convolution can boost the segmentation performance, FBNet equipped with BwBCE and DFM can obtain further improvements (up to $\textbf{2.9\%}$) beyond them. Another minor details should be emphasized that these two modules, i.e., BwBCE and DFM, can facilitate each other ($\textbf{2.9\% \textgreater 0.9\% + 1.2\% }$), which keeps consistent with our discussion in Section \ref{Method}. BwBCE and DFM should be treated as a whole to enhance the feature representation of foreground classes.

\begin{table}[H]
\centering
\setlength{\tabcolsep}{0.8mm}{
\begin{tabular}{m{1.8cm}<{\centering}|m{0.9cm}<{\centering}|m{1.3cm}<{\centering}|m{0.8cm}<{\centering}|m{0.8cm}<{\centering}|m{0.8cm}<{\centering}|m{0.7cm}<{\centering}}
\toprule  
\multirow{2}{*}{Component} & \multicolumn{2}{c|}{\small Auxiliary Loss} & \multicolumn{2}{c|}{FM} & \multirow{2}{*}{mIoU} & \multirow{2}{*}{$\Delta$} \\ \cline{2-5}
                       & P-CE & BwBCE & Conv & DFM &  & \\ \toprule
Baseline               &  - &   -   &   -  &  -  & 77.8 & - \\ \hline
\multirow{3}{*}{}      &   &  \checkmark    &     &    & 78.7 & \textbf{0.9}\textcolor{Red}{\textbf{$\uparrow$}} \\ 
                       &   &     &     &  \checkmark  & 79.0 & \textbf{1.2}\textcolor{Red}{\textbf{$\uparrow$}}  \\ 
                       &  \checkmark  &    &      &    \checkmark  & 80.0 & \textbf{2.2}\textcolor{Red}{\textbf{$\uparrow$}} \\ 
                       &   & \checkmark     &    \checkmark  &     & 79.4 & \textbf{1.6}\textcolor{Red}{\textbf{$\uparrow$}} \\ \hline
FBNet &  -  & \checkmark & -  &  \checkmark &  \textbf{80.7}  & \textbf{2.9}\textcolor{Red}{\textbf{$\uparrow$}} \\
\bottomrule 
\end{tabular}}
\caption{Ablation on each component of FBNet, where Dilated FCN~\cite{DilatedFCN} with the ResNet-101 serves as the baseline. P-CE means point-wise Cross Entropy Loss, Conv means $1\times 1$ convolution, FM means feature modulator.
} 
\label{tab_ab_DFI} 
\end{table}


\subsection{Comparison with state-of-the-art methods}
\noindent\textbf{Results on Cityscapes \textit{test} set.} For a fair and complete comparison with the SOTA methods, we train our models under several different configurations, i.e., only finely annotated data is accessible, extra coarsely annotated data is added, model is initialized with Mapilarry pretrained model. Moreover, we follow the standard settings in \cite{DANet,pspnet,ACFNet,CCNet,CPNet,BiseNet} to adopt multi-scale and flipping testing strategies. As shown in Table \ref{tab_test}, our proposed FBNet achieves $\textbf{82.4\%}$ mIoU with ResNet101 backbone and $\textbf{83.9\%}$ mIoU with WiderResNet38~\cite{widerresnet}, Mapillary~\cite{mapillary} pretrained model and coarse training dataset, which sets up a new SOTA result.
\begin{table}
\centering
\setlength{\tabcolsep}{0.5mm}{
\begin{tabular}{l|c|c|c} 
\toprule  
Method & Conference & Backbone & mIoU(\%) \\
\toprule
PSPNet\cite{pspnet} & CVPR 2017 & ResNet-101 & 78.4 \\
DFN\cite{DilatedFCN} & ICLR 2016& ResNet-101 & 79.3 \\
AAF\cite{AAF} & ECCV 2018 & ResNet-101 & 80.1 \\
DenseASPP\cite{denseaspp} & CVPR 2018 & DenseNet-161 & 80.6 \\
CPNet \cite{CPNet} & CVPR 2020&  ResNet-101 & 81.3 \\
BFP\cite{d_43} & ICCV 2019& ResNet-101 & 81.4 \\
DANet \cite{DANet} & CVPR 2019& ResNet-101 & 81.5 \\
CCNet \cite{CCNet} & ICCV 2019&ResNet-101 & 81.4 \\
SpyGR \cite{SpyGR} & CVPR 2020&ResNet-101 & 81.6 \\
ACFNet \cite{ACFNet} & ICCV 2019&ResNet-101 & 81.8 \\
OCR \cite{OCR} & ECCV 2020&ResNet-101 & 81.8 \\
CDGCNet \cite{CDGCNet} & ECCV 2020&ResNet-101 & 82.0 \\
HANet \cite{HANet} & CVPR 2020 & ResNet-101 & 82.1 \\
RecoNet \cite{RecoNet} & ECCV 2020&ResNet-101 & 82.3 \\
QCO~\cite{QCO} & CVPR 2021& ResNet-101 & 82.3 \\
\midrule
\textbf{Ours}  & N/A& ResNet-101 & \textbf{82.4} \\

\midrule
DeepLabv3+\textsuperscript{\dag} \cite{deeplabv3plus} & ECCV 2018 & Xception& 82.1 \\
SpyGR\textsuperscript{\dag} \cite{SpyGR} & CVPR 2020 & ResNet-101 & 82.3 \\
GSCNN\textsuperscript{\ddag} \cite{GSCNN} & ICCV 2019 & WiderResNet38  & 82.8 \\
HANet\textsuperscript{\dag \ddag} \cite{HANet} & CVPR 2020 & ResNext-101 & 83.2 \\
\midrule
\textbf{Ours}\textsuperscript{\dag \ddag}  & N/A & WiderResNet38  & \textbf{83.9} \\
\bottomrule
\end{tabular}}
\caption{Performance comparisons on the Cityscapes \textit{test} set. $\dag$ means Cityscapes coarsely annotated images are adopted for model training. $\ddag$ means model is initialized by Mapillary pretrained model.} 
\label{tab_test} 
\end{table}


\noindent\textbf{Results on BDD100K.}
To further verify the effect of our proposed method, we evaluate our method on another challenging urban-scene dataset, i.e., BDD100K, based on the DeepLabv3+ \cite{deeplabv3plus}. Table \ref{tab_bdd} shows the detailed results. Obviously, our method also achieves SOTA performance. Comparing with the DeepLabv3+ baseline, FBNet improves the mIoU from $\textbf{64.91\%}$ to $\textbf{66.3\%}$ on the single scale inference without flipping, which demonstrates the effectiveness of our FBNet.


\subsection{Visual inspections and analysis}
\noindent\textbf{Visualization of segmentation results.} 
As shown in Figure \ref{improvemnet_vision}, we take DeepLabv3+~\cite{deeplabv3plus} as baseline to compare segmentation results with ours. It is obvious that FBNet distinguishes more foreground targets from its surrounding, which is highlighted in red dashed boxes in Figure \ref{improvemnet_vision}.

\noindent\textbf{Visualization of FBNet output features.} To further show the validity of our proposed FBNet, we visualize some feature maps generated by FBNet in Figure \ref{attention_vision}.
We can find that foreground targets like poles, traffic lights, trucks, people, motorcycles are all significantly strengthened in the high-level features. Meanwhile, stuff classes are suppressed to some extend. It satisfies our motivation because FBNet is designed as a feature modulator to enhance the deep representation of foreground classes.

\begin{table} [H]
\centering
\setlength{\tabcolsep}{1.6mm}{
 \begin{tabular}{l|c|c|ccc} 
\toprule  
Method & Conference & Backbone & mIoU(\%) \\
\toprule
NiseNet\cite{NiseNet} &ICIP 2019& - & 53.5 \\
FasterSeg\cite{fasterseg} &ICLR 2020& - & 55.3 \\
DeepLabv3+ \cite{deeplabv3plus} &ECCV 2018 & ResNet-101 & 64.9 \\
HANet\cite{HANet} & CVPR 2020&ResNet-101 & 65.6 \\
\midrule
\small\textbf{Ours}  & N/A &ResNet-101 & \textbf{66.3} \\
\bottomrule 
\end{tabular}}
\caption{Performance comparisons on the BDD100K dataset, and all these results are based on the single scale without flipping testing strategy for a fair comparison.} 
\label{tab_bdd} 
 \end{table}

\section{Conclusions}
In this paper, we delve into the essential reason that depresses the foreground targets segmentation performance in urban-scene parsing, which we call \textit{Feature Camouflage}. To relieve the problem, a novel module named Feature Balance Networks(FBNet) is proposed, which is composed of two key components, i.e., Block-wise Binary Cross-Entropy (BwBCE) and Dual Feature Modulator(DFM). BwBCE loss intends to ease feature camouflage from a gradient balanced perspective. At the same time, DFM is designed to strengthen the deep representation of the foreground class in high-level features under the supervision of BwBCE. We conduct extensive experiments to verify the effectiveness of our proposed method, and we achieve a new SOTA result on two challenging urban-scene benchmarks, i.e., Cityscapes and BDD100K.

{\small
\bibliographystyle{ieee_fullname}
\bibliography{fbnet}
}

\end{document}